\pdfoutput=1
\documentclass[letterpaper]{article} 
\usepackage[preprint]{aaai2027} 
\usepackage[hyphens]{url} 
\usepackage{graphicx} 
\usepackage{natbib} 
\usepackage{caption} 
\usepackage{subcaption}
\usepackage{multirow}
\usepackage{amsmath,amssymb}
\usepackage{booktabs}
\title{OrientSAM: Mitigating Camera-Centric Shortcut in Multimodal Spatial Reasoning via Orientation-Aware Spatial Alignment}
\author{
Wenxiao Fan\equalcontrib,
Hang Yin\equalcontrib,
Kan Li
}
\affiliations{
School of Computer Science, Beijing Institute of Technology\\
\{wenxiaofan,yh,likan\}@bit.edu.cn
}

\begin{document}

\maketitle

\begin{abstract}
Multimodal large language models (MLLMs) still struggle with spatial reasoning that requires perspective transformation. In particular, they often rely on camera-centric cues rather than reasoning from the reference object’s viewpoint, leading to systematic errors in non-camera reference settings. In this paper, we first analyze this failure mode and show that object orientation is a key factor underlying such camera-centric shortcut behavior.
To address this issue, we propose \textbf{OrientSAM}, an orientation-aware spatial alignment framework for multimodal models. OrientSAM injects explicit orientation information into multimodal representations through orientation-aware tokens and Fourier-based angle encoding, and further adopts a curriculum learning strategy to progressively improve perspective-aware reasoning. In addition, we build a spatial data construction pipeline to generate orientation-aware spatial supervision from large-scale images.
Experiments on Spatial-MM, ViewSpatial, and 3DSRBench show that OrientSAM consistently outperforms strong baselines, especially on non-camera-view, person-centric, and orientation-sensitive tasks. The results further demonstrate that explicit orientation modeling is important for mitigating camera-centric shortcut behavior and enabling more robust allocentric spatial reasoning in multimodal models.
\end{abstract}

\section{Introduction}

Multimodal large language models (MLLMs) have achieved strong performance on a wide range of vision-language tasks \cite{doi:10.36227/techrxiv.176231405.57942913/v2,DBLP:journals/corr/abs-2509-18905}, yet their spatial reasoning ability remains limited when the task requires perspective transformation rather than direct interpretation from the camera view. In many real-world scenarios, the correct answer is defined by the local viewpoint of a reference object rather than the image plane. For example, judging whether one person is on another person's left or right depends not only on image-plane coordinates, but also on the reference person's facing direction \cite{DBLP:conf/iclr/LinsleyZANGLPS25,DBLP:journals/corr/abs-2510-12276,liu-etal-2023-visual,DBLP:journals/corr/abs-2406-13246}.

This limitation is fundamental for multimodal intelligence in the physical world \cite{DBLP:journals/corr/abs-2509-18905}. Applications such as embodied interaction \cite{DBLP:conf/cvpr/SongBTTSB25}, human-object collaboration \cite{DOGAN2020103654}, and scene understanding \cite{DBLP:conf/iclr/WenJ024,bu-etal-2025-walk} often require reasoning from a reference object's viewpoint rather than the camera's. However, current MLLMs are largely pretrained on camera-view image-text data, making them prone to camera-centric spatial bias.

\begin{figure}[t]
    \centering
    \includegraphics[width=1.0\linewidth]{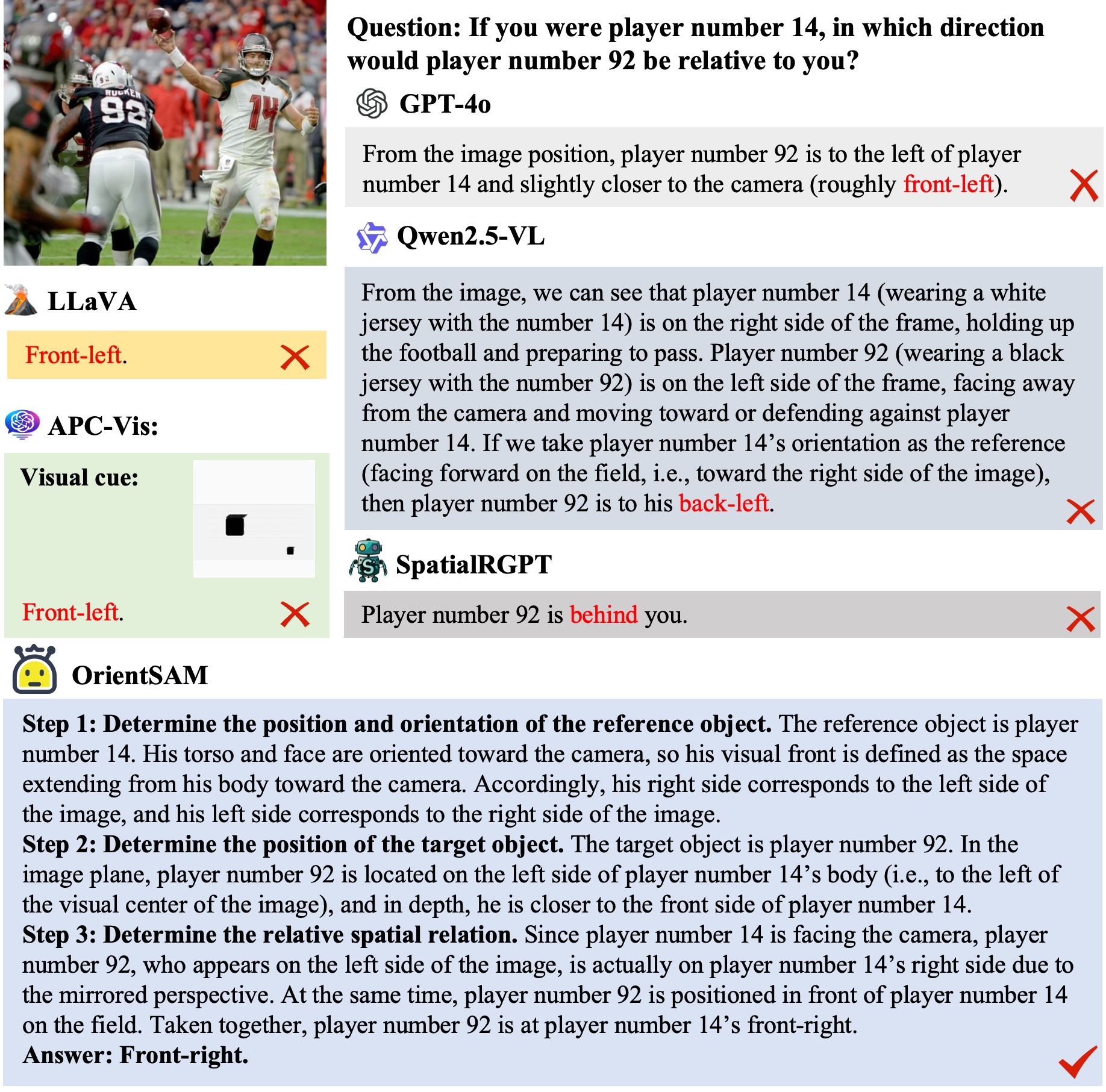}
\caption{A representative case study. Existing models are biased toward camera-plane cues and fail to reason under the reference object’s local viewpoint, whereas OrientSAM correctly predicts the spatial relation by incorporating the reference object’s orientation.}
    \label{fig:case_study}
\end{figure}

Figure~\ref{fig:case_study} shows a representative example. Although the scene is visually clear, several strong MLLMs still fail because their predictions are driven by the left-right arrangement in the image plane, rather than the local reference frame defined by the reference player. Correctly answering such questions requires identifying the reference object's orientation and reasoning about the target position under that viewpoint, suggesting that current MLLMs rely on a simpler camera-centered heuristic instead of true perspective-aware reasoning.

\begin{figure*}[t]
    \centering
    \begin{subfigure}[t]{0.32\linewidth}
        \centering
        \includegraphics[width=\linewidth]{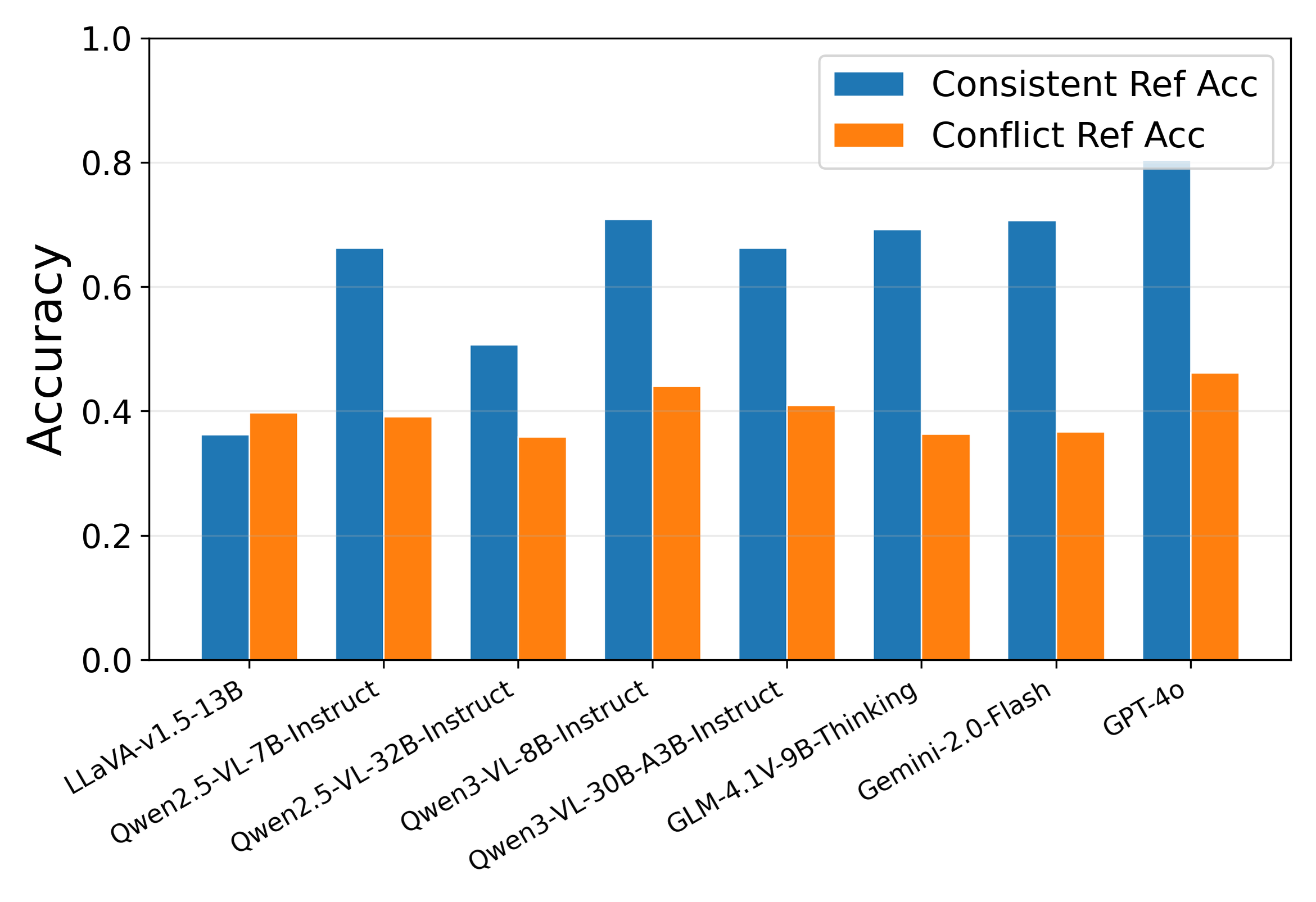}
        \caption{Conflict Is Harder}
    \end{subfigure}
    \hfill
    \begin{subfigure}[t]{0.32\linewidth}
        \centering
        \includegraphics[width=\linewidth]{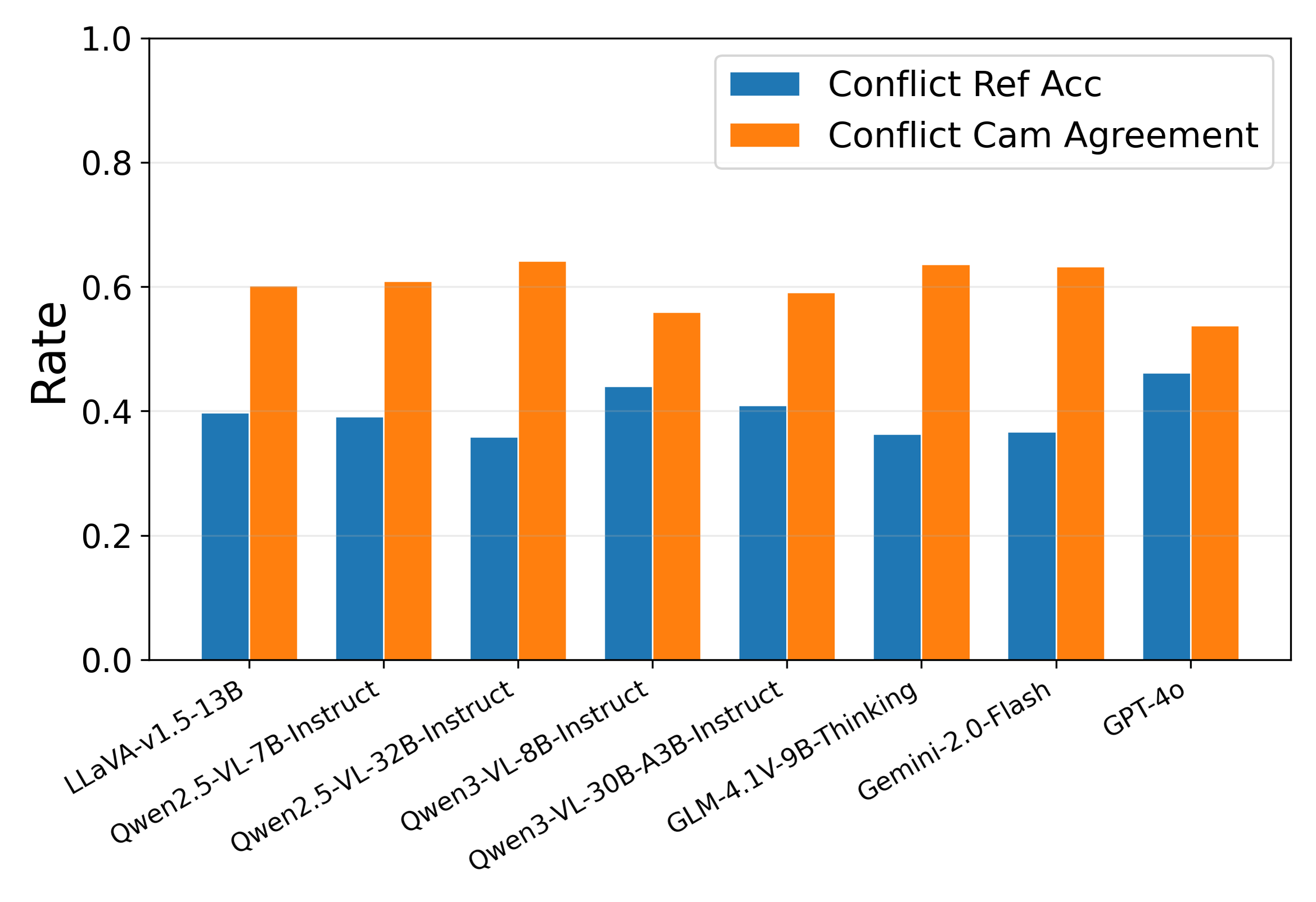}
        \caption{Camera-Centric Agreement Is Higher}
    \end{subfigure}
    \hfill
    \begin{subfigure}[t]{0.32\linewidth}
        \centering
        \includegraphics[width=\linewidth]{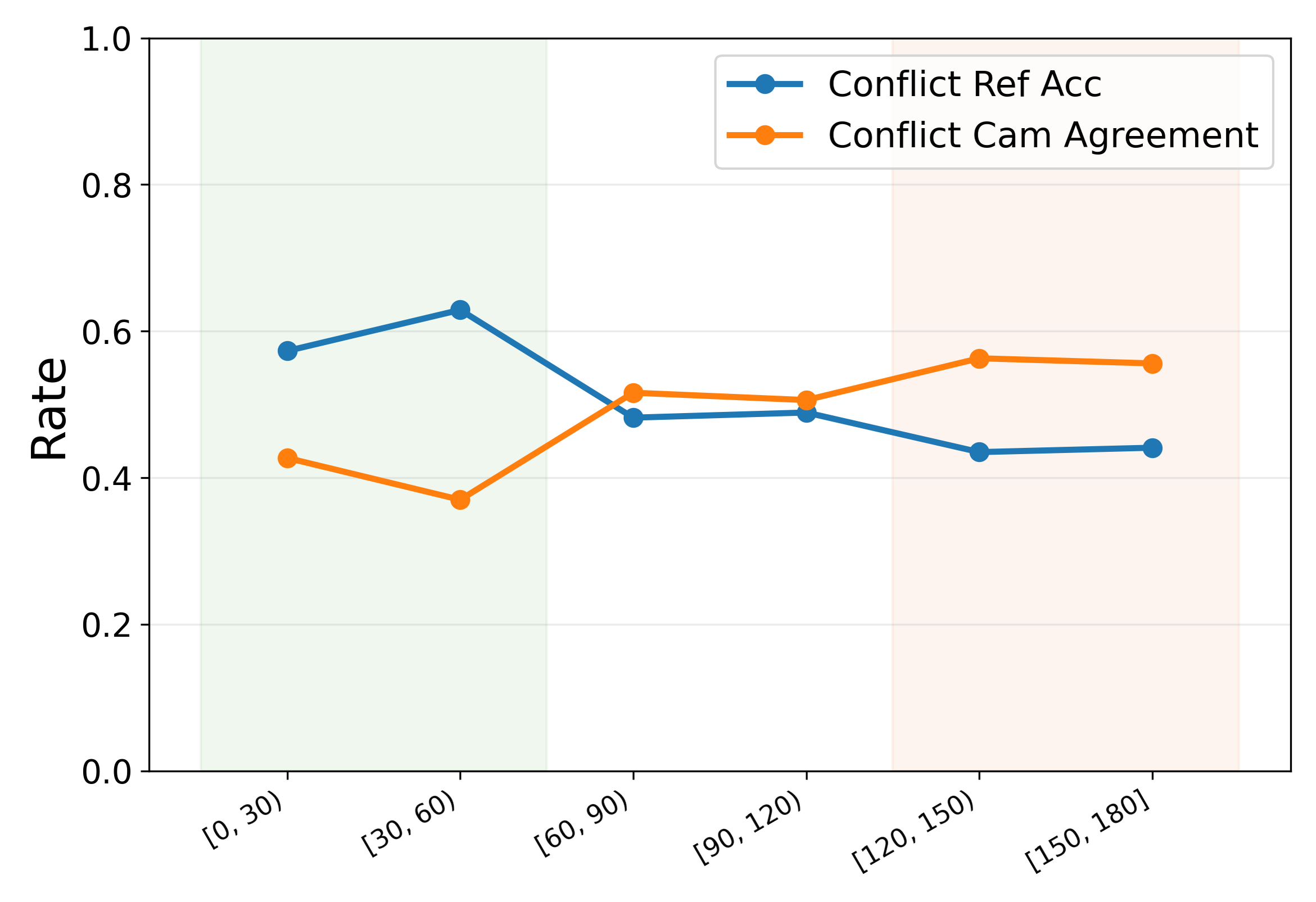}
        \caption{Conflict Angle Curve}
    \end{subfigure}
\caption{
Analysis of camera-centric shortcut behavior on the ViewSpatial benchmark.
(A) \textbf{Conflict Is Harder}: conflict samples are generally more difficult than consistent ones, indicating that models struggle when reference-centric reasoning diverges from image-plane cues.
(B) \textbf{Camera-Centric Agreement Is Higher}: on conflict samples, models show higher agreement with camera-centric answers than with reference-centric ground truth, revealing a clear camera-centric shortcut.
(C) \textbf{Conflict Angle Curve}: the gap between camera-centric agreement and reference-centric accuracy varies across orientation intervals, suggesting that shortcut strength depends on the reference object's orientation.
}
    \label{fig:shortcut}
\end{figure*}

In this paper, we revisit this problem from both diagnostic and modeling perspectives. We introduce a unified analysis framework that disentangles the \emph{reference-centric answer}, defined under the reference object's local frame, from the \emph{camera-centric answer}, defined by the image-plane relation. By comparing model predictions against both, we show that current MLLMs exhibit a clear camera-centric shortcut, especially when the two answers conflict. We further show that this shortcut is closely related to the reference object's orientation, suggesting that the lack of explicit orientation modeling is a key bottleneck for perspective-aware reasoning.

Motivated by this finding, we propose \textbf{OrientSAM}, an \textbf{Orient}ation-aware \textbf{S}patial \textbf{A}lignment \textbf{M}odel for multimodal spatial reasoning. The core idea is to explicitly inject object orientation into multimodal representations and align it with downstream spatial reasoning under non-camera reference frames. Specifically, we build a multi-expert spatial data construction pipeline that integrates region grounding, monocular geometry estimation, and object orientation prediction to automatically construct orientation-aware spatial supervision from large-scale images. Based on this data, OrientSAM introduces orientation-aware tokens together with Fourier-based angle encoding to incorporate orientation priors into the multimodal representation space. We further adopt a curriculum learning strategy that progressively improves the model from basic orientation perception to more challenging perspective-taking reasoning.

Extensive experiments on Spatial-MM, ViewSpatial, and 3DSRBench demonstrate that OrientSAM consistently improves over strong multimodal baselines, with particularly clear gains on non-camera-view, person-centric, and orientation-sensitive tasks. Beyond benchmark improvements, our analysis shows that explicit orientation modeling effectively reduces camera-centric shortcut behavior and leads to more robust reference-centric reasoning.

Our contributions are summarized as follows:
\begin{itemize}
    \item We introduce a unified diagnostic framework for perspective-aware spatial reasoning, and reveal that current MLLMs suffer from a strong camera-centric shortcut, especially when camera-centric and reference-centric answers conflict.
    \item We show that this shortcut is closely related to reference object orientation, and propose OrientSAM, an orientation-aware spatial alignment framework that explicitly incorporates orientation priors into multimodal reasoning.
    \item We build a multi-expert spatial data construction pipeline and a curriculum learning strategy that jointly support the learning of orientation perception and reference-frame transformation.
    \item We demonstrate that OrientSAM consistently outperforms strong baselines on multiple spatial reasoning benchmarks, especially on non-camera-view and orientation-sensitive settings.
\end{itemize}

\begin{table*}[t]
\centering
\setlength{\tabcolsep}{5pt}
\begin{tabular}{l|cccc}
\hline
\textbf{Model} 
& \textbf{Consistent Ref Acc} 
& \textbf{Conflict Ref Acc}  
& \textbf{Conflict Cam Agree}  
& $\Delta_{\text{norm}}$  \\
\hline
LLaVA-v1.5-13B  & 0.362 & 0.398 & 0.602 & +0.513 \\
Qwen2.5-VL-7B-Instruct    & 0.663 & 0.391 & 0.609 & +0.557 \\
Qwen2.5-VL-32B-Instruct  & 0.507 & 0.359 & 0.641 & +0.786 \\
Qwen3-VL-8B-Instruct    & 0.709 & 0.440 & 0.560 & +0.273 \\
Qwen3-VL-30B-A3B-Instruct & 0.662 & 0.409 & 0.591 & +0.445 \\
GLM-4.1V-9B-Thinking    & 0.692 & 0.364 & 0.636 & +0.747 \\
Gemini-2.0-Flash        & 0.707 & 0.367 & 0.633 & +0.725 \\
GPT-4o                  & 0.803 & 0.462 & 0.538 & +0.165 \\
\hline
\textbf{Mean}           & 0.638 & 0.399 & 0.601 & \textbf{+0.526} \\
\hline
\end{tabular}
\caption{
Model-wise analysis of camera-centric shortcut on the ViewSpatial benchmark.
Conflict samples are generally more challenging than consistent ones, indicating the need for perspective transformation.
$\Delta_{\text{norm}} = (\text{CamAgree} - \text{RefAcc}) / \text{RefAcc}$ measures the relative strength of camera-centric bias on conflict samples.
All models exhibit positive $\Delta_{\text{norm}}$, suggesting a consistent tendency to rely on camera-centric cues when camera-centric and reference-centric answers diverge.
}
\label{tab:shortcut_analysis}
\end{table*}

\section{Related Work}

\textbf{Multimodal Spatial Reasoning.} Recent MLLMs have achieved strong performance on general vision-language tasks, yet robust spatial reasoning remains challenging, especially when fine-grained regions, metric distance, or 3D relationships must be inferred. A representative line of work \cite{DBLP:journals/corr/abs-2509-13317,DBLP:journals/corr/abs-2510-18632,DBLP:conf/cvpr/0003XKISGX24,DBLP:journals/corr/abs-2503-19707,ogezi-shi-2025-spare,DBLP:journals/corr/abs-2505-05456} improves spatial reasoning through large-scale spatial data construction and explicit geometric cues. For example, SpatialRGPT \cite{DBLP:conf/nips/ChengYFGYK0L24} builds 3D scene graphs from monocular images and injects region and depth aware supervision to enhance spatial question answering, showing that region grounding and geometric priors are both crucial for spatial understanding. Subsequent work further extends this direction from single images to multi-view settings. SR-3D \cite{DBLP:journals/corr/abs-2509-13317} introduces a unified 2D/3D representation with region prompting and canonical 3D positional embeddings, enabling both single-view and multi-view spatial reasoning within one framework. These methods substantially improve spatial QA, but they mainly focus on general spatial cognition or 3D scene understanding, rather than the more specific challenge of reference-frame transformation under object orientation.

\noindent\textbf{Perspective-taking and Egocentric Bias.}
Another growing line of research studies whether multimodal models can reason from viewpoints other than the camera. Prior analyses \cite{DBLP:journals/corr/abs-2504-17207,DBLP:journals/corr/abs-2601-16378,DBLP:conf/cvpr/0003XKISGX24,DBLP:conf/emnlp/KamathHC23a,DBLP:journals/corr/abs-2409-12969,mayer-etal-2025-ivispar,DBLP:conf/cvpr/YangYGH0X25} show that current VLMs often default to egocentric or image-plane heuristics instead of performing true allocentric reasoning. APC \cite{DBLP:journals/corr/abs-2504-17207} addresses this issue with an external abstraction pipeline that combines object detection, segmentation, depth, and orientation estimation to convert allocentric questions into egocentric ones, demonstrating that perspective-aware reasoning can be improved with explicit scene abstraction. More recently, cognitively inspired token-based approaches \cite{DBLP:journals/corr/abs-2601-16378} show that egocentric bias can also be mitigated by injecting structured perspective cues directly into token space; in particular, embodiment tokens and rotation tokens improve performance on visual perspective-taking benchmarks and suggest that current models may contain latent orientation sensitivity but lack an appropriate representational substrate for viewpoint transformation. Our own analysis further aligns with this trend by showing that, when camera-centric and reference-centric answers conflict, model predictions are consistently more aligned with camera-centric cues, revealing a strong shortcut behavior rather than genuine perspective-taking.

\noindent\textbf{3D-aware Representation and Orientation Modeling.}
To improve spatial reasoning, recent studies \cite{huang2025,DBLP:journals/corr/abs-2504-17207,DBLP:conf/cvpr/JungKKLKC25} increasingly emphasize explicit 3D-aware representation learning and orientation modeling. 3DRS \cite{huang2025} shows that stronger multi-view correspondence is positively correlated with downstream 3D scene understanding, and proposes distilling visual features from pretrained 3D foundation models to enhance 3D awareness in MLLMs. Complementarily, VGGT \cite{DBLP:conf/cvpr/WangCKV0N25} provides a strong general-purpose 3D foundation model that jointly predicts cameras, depth, point maps, and tracks from one or multiple images, offering a versatile source of geometry supervision. Beyond generic 3D geometry, object orientation has recently been identified as a missing but critical factor for spatial reasoning. Orient Anything \cite{DBLP:conf/icml/0001ZPDZ025} demonstrates that even strong VLMs struggle with orientation-sensitive reasoning, and proposes a scalable framework for estimating semantic object orientation from single images. These advances suggest that explicit modeling of both geometry and orientation is necessary for robust spatial understanding. Building on this insight, our work focuses on orientation-aware alignment for multimodal reasoning, targeting the reference-frame transformation problem that is insufficiently addressed by prior depth-only or generic 3D-aware methods.

\begin{figure*}[t]
    \centering
    \includegraphics[width=0.75\linewidth]{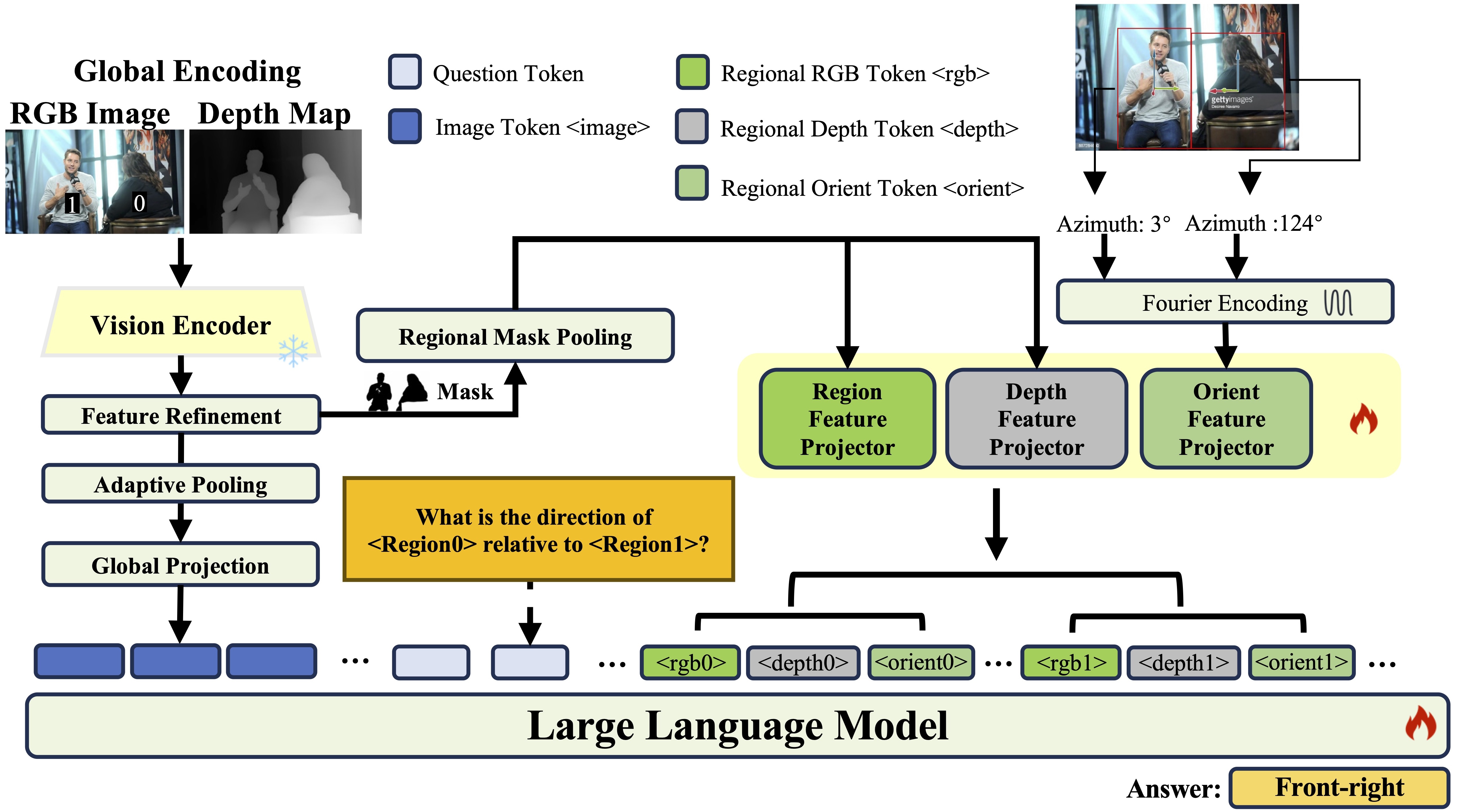}
    \caption{Overview of OrientSAM. The model augments multimodal reasoning with region-level RGB, depth, and orientation tokens. Object orientation is encoded with Fourier features and injected into the language model to support reference-centric spatial reasoning.}
    \label{fig:placeholder1}
\end{figure*}

\section{Camera-Centric Shortcut in VLMs}




To better understand the limitations of current multimodal large language models (MLLMs) in perspective-aware spatial reasoning, we conduct a systematic analysis of whether these models genuinely reason from a reference object's viewpoint, or instead rely on a simpler camera-centric heuristic.


\noindent\textbf{Problem Setup.}
We study perspective-taking tasks that ask for the spatial relation between two objects from the viewpoint of a reference object. Solving such tasks requires transforming the observed scene into the reference object's local frame, rather than directly relying on image-plane layout.

To distinguish these two behaviors, we define two answers for each sample. The \emph{reference-centric answer}, $\mathrm{Ans}_{\text{ref}}$, is the ground-truth relation under the reference object's viewpoint. The \emph{camera-centric answer}, $\mathrm{Ans}_{\text{cam}}$, is the relation implied by the horizontal image-plane configuration. Based on whether they agree, we split the data into two subsets: a \emph{consistent} subset, where $\mathrm{Ans}_{\text{ref}} = \mathrm{Ans}_{\text{cam}}$, and a \emph{conflict} subset, where $\mathrm{Ans}_{\text{ref}} \neq \mathrm{Ans}_{\text{cam}}$. This split allows us to test whether a model truly performs reference-centric reasoning or simply follows camera-view cues.

\noindent\textbf{Metrics.}
We evaluate model behavior from two complementary perspectives.
First, we compute \emph{reference-centric accuracy}, i.e., whether the prediction matches $\mathrm{Ans}_{\text{ref}}$.
Second, we compute \emph{camera-centric agreement}, i.e., whether the prediction matches $\mathrm{Ans}_{\text{cam}}$.
Importantly, accuracy is always measured against $\mathrm{Ans}_{\text{ref}}$, while $\mathrm{Ans}_{\text{cam}}$ is used only for behavioral diagnosis.
This distinction allows us to separate true reasoning performance from shortcut consistency.

\noindent\textbf{Conflict samples are generally harder.}
Figure~\ref{fig:shortcut}A compares model performance on the consistent and conflict subsets of ViewSpatial.
Across most evaluated models, conflict samples are substantially more difficult than consistent ones.
This result suggests that current MLLMs do not fail uniformly on spatial reasoning tasks; rather, their failures become much more pronounced precisely when image-plane cues and reference-centric reasoning diverge.
In other words, the challenge is not merely recognizing object positions, but correctly transforming them into the reference object's viewpoint.

\noindent\textbf{Models prefer camera-centric answers under conflict.}
Figure~\ref{fig:shortcut}B and Table~\ref{tab:shortcut_analysis} further reveal the key shortcut pattern.
On the conflict subset, camera-centric agreement is consistently higher than reference-centric accuracy for all evaluated models, indicating that model predictions are more aligned with camera-view relations than with the true reference-centric answer.
This pattern is especially important because the conflict subset removes the ambiguity between the two strategies: once $\mathrm{Ans}_{\text{ref}}$ and $\mathrm{Ans}_{\text{cam}}$ disagree, elevated agreement with $\mathrm{Ans}_{\text{cam}}$ directly signals shortcut behavior rather than genuine perspective-taking.

To quantify this tendency, we report $\Delta_{\text{norm}} = \frac{\text{CamAgree} - \text{RefAcc}}{\text{RefAcc}}$,
which measures the relative strength of camera-centric bias on conflict samples.
As shown in Table~\ref{tab:shortcut_analysis}, all models exhibit positive $\Delta_{\text{norm}}$, further confirming that the shortcut is systematic rather than anecdotal.

\noindent\textbf{Shortcut strength depends on reference orientation.}
Beyond the conflict/consistent split, we further ask when this shortcut becomes stronger.
Figure~\ref{fig:shortcut}C shows that the gap between camera-centric agreement and reference-centric accuracy varies across orientation intervals of the reference object.
Rather than suggesting a strictly monotonic trend, this result indicates a clear orientation-dependent contrast: model behavior changes noticeably as the reference object's facing direction becomes more or less aligned with the image plane.
This observation suggests that the shortcut is closely tied to whether the model can correctly encode and use the reference object's orientation during viewpoint transformation.

Overall, these results reveal that current MLLMs exhibit a strong camera-centric shortcut in perspective-aware spatial reasoning.
More importantly, our analysis shows that this failure mode is not simply a generic weakness in spatial QA, but is tightly linked to reference-frame transformation and the handling of object orientation.

\noindent\textbf{Implication for modeling.}
The above findings suggest that improving perspective-aware spatial reasoning requires more than stronger visual perception or additional generic geometric cues.
The main bottleneck lies in the absence of an explicit mechanism for representing the reference object's orientation and using it to transform spatial relations into a local reference frame.
Therefore, an effective solution should satisfy two requirements:
(1) it should inject object orientation into the multimodal representation in an explicit and learnable form; and
(2) it should align such orientation cues with downstream spatial reasoning, especially on samples where camera-centric and reference-centric answers conflict.
Motivated by this, we propose OrientSAM, an orientation-aware spatial alignment framework that explicitly incorporates orientation priors into multimodal representations and progressively improves perspective-taking ability through alignment learning and curriculum training.

\section{Orientation-aware Spatial Alignment}



Motivated by the findings in Section 3, we observe that current MLLMs fail to perform reliable perspective-taking due to the lack of explicit orientation modeling, leading to a strong reliance on camera-centric shortcuts. To address this limitation, we propose \textbf{Orientation-aware Spatial Alignment (OrientSAM)}, a framework that explicitly incorporates object orientation into multimodal representations and aligns it with spatial reasoning.

\noindent\textbf{Problem Formulation.}
We consider a perspective-taking task where the goal is to determine the spatial relation between a query object $o_q$ and a reference object $o_r$ under the reference object's coordinate system.

Given an input image $I$, we extract a set of object regions $\{o_i\}_{i=1}^N$, where each object is associated with a bounding box and visual features. For each object $o_i$, we further associate an orientation representation $\mathbf{\omega}_i$, which encodes its semantic facing direction in 3D space.

The task is to predict the relative spatial relation:
\begin{equation}
y = f(I, o_r, o_q, \mathbf{\omega}_r),
\end{equation}
where $y$ is defined under the reference-centric coordinate system induced by the orientation of $o_r$.

Unlike conventional approaches that implicitly rely on image-plane coordinates, our formulation explicitly models the transformation induced by object orientation, enabling allocentric reasoning.

\noindent\textbf{Multi-expert Spatial Data Construction.}
To enable orientation-aware learning, we construct a multi-expert data pipeline that augments 2D images with geometric and orientation annotations. The pipeline integrates several expert models to recover spatial attributes from monocular images.

Given an image $I$, we first extract a set of object instances:
\begin{equation}
O = \{o_i\}_{i=1}^N,
\end{equation}
where each instance is represented as:
\begin{equation}
o_i = (b_i, m_i, t_i),
\end{equation}
with bounding box $b_i$, segmentation mask $m_i$, and semantic label $t_i$.

We then estimate the 3D structure of the scene using a monocular geometry model:
\begin{equation}
(D, K, \xi) = F_{\text{geo}}(I),
\end{equation}
where $D$ denotes the depth map, $K$ the camera intrinsic parameters, and $\xi$ the camera pose.

The depth of each object is computed as the median depth within its mask:
\begin{equation}
z_i = \mathrm{Median}\{D(u,v) \mid (u,v) \in m_i\}.
\end{equation}

To capture object orientation, we employ an orientation estimation model to obtain angular representations:
\begin{equation}
(\theta_i, \phi_i, \delta_i, c_i) = F_{\text{orient}}(I, b_i),
\end{equation}
where $\theta_i$, $\phi_i$, and $\delta_i$ denote polar, azimuth, and rotation angles, respectively, and $c_i$ is the confidence score.

Each object is thus represented as a unified attribute vector:
\begin{equation}
\mathbf{v}_i = \langle t_i, b_i, m_i, z_i, \theta_i, \phi_i, \delta_i, c_i \rangle.
\end{equation}

Based on these attributes, we automatically construct spatial reasoning samples by selecting object pairs and generating reference-centric questions and answers. This process provides large-scale supervision signals for learning orientation-aware reasoning.

In practice, the automatically constructed supervision is further filtered by data quality control and organized in a staged manner. 
For orientation perception, we retain only high-confidence samples whose orientation predictions are sufficiently reliable, so that the injected \texttt{<orient>} tokens are anchored to stable semantic directions. 
For perspective-aware reasoning, we construct multi-object question-answer pairs under reference-centric coordinate systems and combine them with a moderate amount of general instruction data to reduce catastrophic forgetting. 
Following this design, the training data are organized into two stages: an orientation perception stage for learning stable orientation primitives, and a relation reasoning stage for learning reference-frame transformation and multi-object spatial reasoning. 
This staged data construction improves both the reliability of supervision and the stability of downstream alignment learning. 

\noindent\textbf{Orientation-aware Representation}
As discussed in Section~3, a major limitation of current MLLMs is that they mainly rely on image-plane layout while lacking an explicit representation of object orientation. As a result, they often fail to establish the correct local reference frame for perspective-aware reasoning. To address this issue, we explicitly encode object orientation and inject it into the multimodal representation space.

For each object instance $o_i$, we already obtain its geometric-semantic attribute vector in Sec.~4.2:
\begin{equation}
\mathbf{v}_i=\langle t_i,b_i,m_i,z_i,\theta_i,\phi_i,\delta_i,c_i\rangle,
\end{equation}
where $\phi_i \in [0,2\pi)$ denotes the azimuth angle that characterizes the semantic facing direction of the object.

A direct scalar representation of $\phi_i$ is problematic because angular variables are periodic. In particular, two angles close to $0$ and $2\pi$ correspond to nearly identical orientations but are far apart in Euclidean space. To preserve this circular structure, we adopt Fourier feature encoding:
\begin{equation}
\gamma(\phi_i)=\left[\sin(2^k\phi_i),\cos(2^k\phi_i)\right]_{k=0}^{K-1},
\end{equation}
where $K$ denotes the frequency order.



Let $\mathbf{r}_i \in \mathbb{R}^{d_r}$ denote the region-level RGB feature of object $o_i$, obtained by mask pooling over the visual feature map within its segmented region, and let $\mathbf{d}_i \in \mathbb{R}^{d_d}$ denote the corresponding depth feature.
We project the orientation encoding into the same embedding space and inject it into the visual representation:
\begin{equation}
\tilde{\mathbf{r}}_i = \mathbf{r}_i + W_o \gamma(\phi_i),
\end{equation}
where $W_o$ is a learnable projection matrix.

On the language side, we also augment the object token associated with $o_i$ using the same orientation encoding:
\begin{equation}
\tilde{\mathbf{e}}_i = \mathrm{Embed}(t_i) + W_t \gamma(\phi_i),
\end{equation}
where $t_i$ is the semantic label or region identifier of object $o_i$, and $W_t$ is another learnable projection matrix.

The final multimodal representation is then constructed by jointly encoding the orientation-enhanced visual and textual features:
\begin{equation}
\mathcal{H} = \mathrm{MMEncoder}\big(\{\tilde{\mathbf{r}}_i,\mathbf{d}_i\}_{i=1}^{N},\{\tilde{\mathbf{e}}_i\}_{i=1}^{N}\big).
\end{equation}

In this way, object orientation is explicitly represented as part of the multimodal context, rather than being treated as an auxiliary textual hint. This formulation allows the model to jointly reason over object appearance, depth, and semantic facing direction, which is essential for constructing reference-centric spatial relations.

\begin{table*}[t]
\centering
\setlength{\tabcolsep}{4pt}
\begin{tabular}{l|cc|cccc}
\hline
\multirow{2}{*}{\textbf{Model}} & \multicolumn{2}{c|}{\textbf{Spatial-MM}} & \multicolumn{4}{c}{\textbf{ViewSpatial}} \\
\cline{2-7}
& \textbf{Cam} & \textbf{Non-Cam} & \textbf{Cam-Obj} & \textbf{Cam-Rel} & \textbf{P-Obj} & \textbf{P-Rel} \\
\hline
Random Guess & 37.61 & 39.93 & 25.00 & 25.00 & 25.00 & 30.14 \\
\hline
LLaVA-v1.5-7B & 41.34 & 40.30 & 26.50 & 27.24 & 24.00 & 27.91 \\
LLaVA-v1.5-13B & 51.24 & 46.27 & 25.60 & 28.14 & 23.39 & 28.15 \\
Qwen2.5-VL-3B & 64.66 & 32.09 & 30.22 & 40.67 & 38.55 & 30.76 \\
Qwen2.5-VL-7B & 77.21 & 37.31 & 26.00 & 44.61 & 37.65 & 32.66 \\
Qwen2.5-VL-32B & 78.80 & 38.81 & 31.53 & 57.66 & 56.85 & 39.07 \\
Qwen3-VL-2B & 77.56 & 35.82 & 23.80 & 48.11 & 44.98 & 38.84 \\
Qwen3-VL-4B & 82.51 & 32.09 & 31.22 & 49.63 & 43.07 & 40.50 \\
Qwen3-VL-8B & 82.69 & 38.81 & 31.83 & 53.13 & 51.57 & 39.55 \\
\hline
SpatialRGPT & 89.75 & 39.55 & 69.28 & \underline{78.62} & 69.18 & 41.33 \\
APC-Vis & \textbf{92.23} & \underline{86.57} & \underline{79.62} & 73.32 & \textbf{75.40} & \underline{72.57} \\
\hline
Gemini-2.0-Flash & 77.91 & 27.61 & 28.57 & 46.19 & 43.37 & 38.60 \\
GPT-4o & 75.62 & 41.79 & 37.47 & 40.16 & 44.18 & 44.18 \\
\hline
OrientSAM & \underline{89.40} & \textbf{87.31} & \textbf{83.33} & \textbf{80.49} & \underline{73.49} & \textbf{85.04} \\
\hline
\end{tabular}
\caption{Performance comparison on Spatial-MM and ViewSpatial (Accuracy, \%). Cam and Non-Cam denote camera-view and non-camera-view; Obj and Rel denote object orientation and relative direction; P denotes person-view. Bold and underline indicate the best and second-best results.}
\label{tab:main_spatial_view}
\end{table*}

\begin{table}[t]
\centering
\setlength{\tabcolsep}{2pt}
\begin{tabular}{l|cccc}
\hline
\textbf{Model} & \textbf{Height} & \textbf{Loc.} & \textbf{Orient.} & \textbf{Multi.} \\
\hline
Random Guess      & 50.00 & 50.00 & 37.56 & 41.69 \\
\hline
LLaVA-v1.5-7B     & 50.14 & 64.37 & 42.17 & 45.16 \\
LLaVA-v1.5-13B    & 57.25 & 61.26 & 39.16 & 42.66 \\
LLaVA-NeXT-7B     & 54.20 & 57.87 & 39.45 & 42.83 \\
Qwen2.5-VL-3B     & 51.59 & 58.86 & 41.59 & 43.87 \\
Qwen2.5-VL-7B    & 57.97 & 65.12 & 50.77 & 52.59 \\
Qwen2.5-VL-32B   & 52.17 & 62.84 & 41.12 & 41.87 \\
Qwen3-VL-2B      & 50.94 & 66.94 & 43.62 & 44.26 \\
Qwen3-VL-4B      & 52.39 & 73.87 & 48.15 & 50.02 \\
Qwen3-VL-8B      & 54.78 & 73.93 & 49.09 & 50.22 \\
\hline
SpatialRGPT     & 55.36 & \underline{77.15} & 65.37 & 50.46 \\
APC-Vis       & 54.49 & 56.24 & \underline{79.41} & 47.04 \\
\hline
Gemini-2.0-Flash  & \underline{63.91} & 58.22 & 35.14 & 57.80 \\
GPT-4o         & \textbf{74.35} & 67.00 & 35.24 & \underline{58.86} \\
\hline
OrientSAM         & 56.67 & \textbf{78.23} & \textbf{81.83} & \textbf{77.80} \\
\hline
\end{tabular}
\caption{Performance comparison on 3DSRBench.  ''Loc.'' denotes Location and Distance, ''Orient.'' denotes Orientation and Direction, and ''Multi.'' denotes Multi-object Relation.}
\label{tab:main_3dsrbench}
\end{table}

\noindent\textbf{Orientation-aware Alignment Learning}
Although the representation in Sec.~4.3 introduces explicit orientation information, such information cannot be effectively used unless it is properly aligned with the semantic space of the pretrained MLLM. In practice, directly appending geometric signals to the input often yields limited gains, because pretrained models are strongly biased toward RGB-dominant visual patterns and may ignore newly introduced spatial cues. Therefore, beyond representation design, a key challenge is to align RGB, depth, and orientation information within a unified reasoning space.

To this end, we project the three modalities associated with each object instance $o_i$ into the language embedding space through independent projectors:
\begin{equation}
\mathbf{e}_i^{\text{rgb}} = W_r \tilde{\mathbf{r}}_i,\qquad
\mathbf{e}_i^{\text{depth}} = W_d \mathbf{d}_i,\qquad
\mathbf{e}_i^{\text{orient}} = W_{\phi}\gamma(\phi_i),
\end{equation}
where $W_r$, $W_d$, and $W_{\phi}$ are learnable projection matrices.

Instead of collapsing all geometric cues into a single feature vector, we preserve their semantic roles by organizing them as structured region-level tokens:
\begin{equation}
\mathcal{T}_i=\{\langle rgb_i\rangle,\langle depth_i\rangle,\langle orient_i\rangle\}.
\end{equation}
This design allows the language model to dynamically attend to different sources of information depending on the reasoning demand. This disentangled design is preferable to directly merging all spatial cues into a single token, since it preserves modality-specific semantics and makes the contribution of appearance, depth, and orientation easier to access during reasoning.In particular, $\langle rgb_i\rangle$ provides appearance cues, $\langle depth_i\rangle$ supports front-back disambiguation, and $\langle orient_i\rangle$ specifies the local coordinate system induced by the reference object.

Given an image-question pair, the final input sequence is constructed as
\begin{equation}
\mathcal{X}=\{\langle image\rangle,\mathcal{T}_1,\mathcal{T}_2,\dots,\mathcal{T}_N,\text{question tokens}\}.
\end{equation}
Based on this aligned multimodal input, the model predicts the reference-centric answer $y^{\text{ref}}$ using the standard next-token prediction objective:
\begin{equation}
\mathcal{L}_{\text{NTP}}=-\log p_{\Theta}\left(y^{\text{ref}} \mid \mathcal{X}\right),
\end{equation}
where $\Theta$ denotes all trainable parameters.
Notably, we do not explicitly introduce an auxiliary loss against the camera-centric answer. Instead, shortcut suppression is achieved implicitly through multimodal alignment: once the model is forced to condition its prediction on the aligned RGB-depth-orientation token structure, it becomes less likely to rely solely on image-plane cues and more capable of performing reference-centric reasoning. In this sense, orientation-aware alignment serves as the bridge between low-level geometric priors and high-level language reasoning.

\noindent\textbf{Curriculum Learning Strategy}
Learning perspective-aware spatial reasoning is inherently challenging because the model must first acquire reliable orientation perception and then use it to perform reference-frame transformation. Directly optimizing complex perspective-taking tasks from the beginning often leads to unstable training and shortcut solutions, where the model memorizes superficial image-plane correlations rather than learning object-centric reasoning. To address this issue, we adopt a curriculum learning strategy that gradually increases task difficulty.

\textbf{Stage 1: Orientation Perception.}
In the first stage, the model is trained to recognize the orientation of individual objects. Given an object instance $o_i$, the task is to predict its azimuth angle $\phi_i$:
\begin{equation}
\min_{\Theta}\; \mathbb{E}\big[\ell_{\text{orient}}(\hat{\phi}_i,\phi_i)\big],
\end{equation}
where $\hat{\phi}_i$ is the predicted azimuth and $\ell_{\text{orient}}$ denotes the orientation supervision loss. This stage encourages the model to establish a stable mapping from regional appearance to semantic facing direction, thereby learning reliable orientation-aware embeddings.

\begin{table*}[t]
\centering
\begin{tabular}{l|cc|cc|cc|c}
\hline
\multirow{2}{*}{\textbf{Model}} 
& \multicolumn{2}{c|}{\textbf{Comp.}} 
& \multicolumn{2}{c|}{\textbf{Spatial-MM}} 
& \multicolumn{2}{c|}{\textbf{ViewSpatial}} 
& \textbf{3DSR} \\
\cline{2-8}
& \textbf{Orient} & \textbf{CL} 
& \textbf{Cam} & \textbf{Non-Cam} 
& \textbf{P-Obj} & \textbf{P-Rel} 
& \textbf{Ori.-Dir.} \\
\hline
Base-VLM (Zero-shot) & $\times$ & $\times$ & 77.15 & 33.40 & 32.13 & 34.59 & 41.21 \\
Base-VLM (SFT)       & $\times$ & $\times$ & 87.65 & 43.15 & 44.30 & 42.12 & 50.35 \\
OrientSAM (w/o CL)   & $\checkmark$ & $\times$ & 88.54 & 76.29 & 65.40 & 74.32 & 72.83 \\
OrientSAM (Full)     & $\checkmark$ & $\checkmark$ & \textbf{89.40} & \textbf{87.31} & \textbf{73.49} & \textbf{85.04} & \textbf{81.83} \\
\hline
\end{tabular}
\caption{Ablation results of orientation feature injection and curriculum learning (Accuracy, \%). Comp. denotes components; Cam and Non-Cam denote camera-view and non-camera-view, respectively; P-Obj and P-Rel denote person-view object orientation and person-view relative direction, respectively; 3DSR denotes 3DSRBench, and Ori.-Dir. denotes orientation and direction.}
\label{tab:ablation_main}
\end{table*}

\textbf{Stage 2: Orientation-aware Reasoning.}
After the model has acquired stable orientation perception, training proceeds to full perspective-aware spatial reasoning. In this stage, the model receives the structured RGB-depth-orientation token sequence defined above and is optimized to predict the correct reference-centric relation:
\begin{equation}
\min_{\Theta}\; \mathcal{L}_{\text{NTP}}.
\end{equation}

Unlike Stage~1, this objective requires the model not only to identify object orientation, but also to use it to transform spatial relations into the local reference frame of the reference object.

This progressive training strategy offers two advantages. First, it decomposes a difficult reasoning problem into perception and reasoning subproblems, making optimization more stable. Second, it prevents the model from prematurely converging to camera-centric shortcuts, since the model is first encouraged to build explicit sensitivity to orientation before being asked to solve full reference-frame transformation tasks. Through this curriculum, the model gradually evolves from basic orientation understanding to robust reference-centric spatial reasoning under varying viewpoint conditions.

\begin{table*}[t]
\centering
\setlength{\tabcolsep}{7pt}
\renewcommand{\arraystretch}{1.08}
\begin{tabular}{l|cc|cc|c}
\hline
\multirow{2}{*}{\textbf{Encoding}} 
& \multicolumn{2}{c|}{\textbf{Spatial-MM}} 
& \multicolumn{2}{c|}{\textbf{ViewSpatial}} 
& \textbf{3DSRBench} \\
\cline{2-6}
& \textbf{Camera-view} & \textbf{Non-camera-view} 
& \textbf{Person-Obj Ori.} & \textbf{Person-Rel Dir.} 
& \textbf{Ori. \& Dir.} \\
\hline
Scalar & 86.53 & 75.26 & 62.41 & 72.13 & 73.52 \\
Fourier ($N=1$) & 88.21 & 83.09 & 69.53 & 80.37 & 78.44 \\
Fourier ($N=2$) & \textbf{89.40} & \textbf{87.31} & \textbf{73.49} & \textbf{85.04} & \textbf{81.83} \\
Fourier ($N=3$) & 89.13 & 86.91 & 72.82 & 84.63 & 81.17 \\
\hline
\end{tabular}
\caption{Ablation study of different angle encoding strategies (Accuracy, \%).}
\label{tab:ablation_encoding}
\end{table*}
\section{Experiments}

\noindent\textbf{Experimental Setup.}
We evaluate OrientSAM on three spatial reasoning benchmarks: \textbf{Spatial-MM} \cite{DBLP:conf/emnlp/ShiriGF0HL24}, \textbf{3DSRBench} \cite{DBLP:journals/corr/abs-2412-07825}, and \textbf{ViewSpatial} \cite{DBLP:journals/corr/abs-2505-21500}. 
Spatial-MM contains both camera-centric and non-camera-centric spatial question answering pairs. Following the setting in our chapter experiments, we use 566 camera-view question-answer pairs and 132 non-camera-view pairs for evaluation. 
3DSRBench focuses on deeper 3D spatial reasoning from single images, and we follow its four subtask partition, including \textit{Height} (690 samples), \textit{Location} (1709 samples), \textit{Orientation} (1036 samples), and \textit{Multi-object Relation} (1722 samples). 
ViewSpatial is used to evaluate viewpoint transformation and spatial consistency under multi-view settings, with four evaluation dimensions: camera-view object orientation, camera-view relative direction, person-view object orientation, and person-view relative direction. 

We compare OrientSAM with representative open-source multimodal models, including LLaVA-v1.5-7B \cite{DBLP:conf/cvpr/LiuLLL24}, LLaVA-v1.5-13B \cite{DBLP:conf/cvpr/LiuLLL24}, LLaVA-NeXT-7B \cite{liu2024llavanext}, Qwen2.5-VL-3B/7B/32B \cite{qwen2.5-VL}, and Qwen3-VL-2B/4B/8B \cite{qwen3technicalreport}, as well as strong spatial reasoning models such as SpatialRGPT \cite{DBLP:conf/nips/ChengYFGYK0L24} and APC-Vis \cite{DBLP:journals/corr/abs-2504-17207}. 
We further include proprietary models Gemini-2.0-Flash \cite{google_gemini20flash_2025} and GPT-4o \cite{DBLP:journals/corr/abs-2303-08774} for reference.

\subsection{Main Results}

\noindent\textbf{Results on Spatial-MM and ViewSpatial.}
Table~\ref{tab:main_spatial_view} reports the comparison on Spatial-MM and ViewSpatial. Overall, OrientSAM performs strongest on the subsets that require explicit reference-frame transformation and orientation-aware reasoning. On Spatial-MM, our method achieves 87.31\% on the non-camera-view setting, outperforming all compared baselines, which shows that explicit orientation modeling substantially improves reasoning under non-camera reference frames. On the camera-view setting, although APC-Vis attains the highest score, OrientSAM still achieves a competitive 89.40\%, indicating that the proposed orientation-aware alignment improves perspective-sensitive reasoning without sacrificing standard camera-view performance.

On ViewSpatial, OrientSAM achieves the best performance on three out of four subtasks, including camera-view object orientation, camera-view relative direction, and person-view relative direction. The gain is particularly clear on the person-view relative direction task, where our method surpasses APC-Vis by a large margin (85.04\% vs. 72.57\%), demonstrating stronger robustness when the model must move beyond the camera coordinate system and reason under a person-centric frame. For person-view object orientation, OrientSAM is slightly below APC-Vis, but still substantially stronger than general-purpose MLLMs. Taken together, these results suggest that the main advantage of OrientSAM lies in explicitly enhancing orientation-conditioned reasoning in the settings where camera-view cues are insufficient or misleading.

\noindent\textbf{Results on 3DSRBench.}
We further compare different models on 3DSRBench, and the results are shown in Table~\ref{tab:main_3dsrbench}. Compared with general-purpose MLLMs, OrientSAM achieves clear advantages on the more spatially intensive subtasks. In particular, our method obtains the best results on Location and Distance, Orientation and Direction, and Multi-object Relation. The gain on Orientation and Direction is especially notable, where OrientSAM outperforms both APC-Vis and SpatialRGPT, confirming the effectiveness of explicit orientation-aware modeling. On Multi-object Relation, OrientSAM also substantially surpasses all compared models, indicating that it is better able to maintain geometric consistency under multiple entities and relational constraints.

However, in the Height subtask, OrientSAM does not achieve the best performance. GPT-4o and Gemini-2.0-Flash remain stronger on this category. This suggests that height comparison may depend more on general visual perception and common sense calibration, whereas the main advantage of OrientSAM is concentrated on direction-sensitive and reference-dependent reasoning. Overall, the results on 3DSRBench further validate that OrientSAM is particularly effective for tasks involving object orientation, relative direction, and multi-object spatial constraints, which are exactly the scenarios where conventional MLLMs tend to exhibit strong camera-centric bias.

\subsection{Ablation Study}

\textbf{Effect of Orientation Feature Injection and Curriculum Learning.}
Table~\ref{tab:ablation_main} reports the ablation results of orientation injection and curriculum learning. We compare four settings: Base-VLM (Zero-shot), Base-VLM (SFT), OrientSAM (w/o CL), and OrientSAM (Full).
Three observations are clear. 
First, supervised fine-tuning alone brings only limited gains on reference-dependent tasks. Although Base-VLM (SFT) improves over the zero-shot baseline, its performance remains low on non-camera-view and person-centric reasoning, e.g., 43.15\% on Spatial-MM non-camera-view and 42.10\% on ViewSpatial person-relative direction. This indicates that data fine-tuning alone is insufficient for robust perspective transformation.
Second, explicit orientation injection is the main source of performance gain. After introducing the orientation branch, OrientSAM (w/o CL) substantially improves over Base-VLM (SFT) across all orientation-sensitive tasks, including Spatial-MM non-camera-view (76.20\% vs.~43.15\%), ViewSpatial person-relative direction (74.30\% vs.~42.10\%), and 3DSRBench orientation \& direction (72.80\% vs.~50.35\%). This confirms the importance of explicit orientation-aware representation.
Third, curriculum learning further improves multimodal alignment and unlocks the full potential of the model. Compared with OrientSAM (w/o CL), the full model further improves Spatial-MM non-camera-view from 76.20\% to 87.31\%, ViewSpatial person-relative direction from 74.30\% to 85.04\%, and 3DSRBench orientation \& direction from 72.80\% to 81.83\%. These gains suggest that progressive training from orientation perception to perspective-aware reasoning is more effective than directly optimizing all tasks in a single stage.


\textbf{Effect of Angle Encoding Strategy.}
Table~\ref{tab:ablation_encoding} compares different angle encoding strategies. Direct scalar encoding gives the weakest results, especially on non-camera-view and person-centric tasks, showing that naive scalar regression cannot properly model the periodic nature of orientation. Replacing scalar encoding with Fourier features consistently improves performance across all benchmarks. Among all variants, Fourier ($N=2$) achieves the best overall results, including 87.31\% on Spatial-MM non-camera-view, 85.04\% on ViewSpatial person-relative direction, and 81.83\% on 3DSRBench orientation \& direction. Increasing the frequency order to $N=3$ does not bring further gains, suggesting that second-order Fourier encoding provides the best balance between periodic continuity and discriminability.

\subsection{Orientation-based Analysis}

\begin{table}[t]
\centering
\setlength{\tabcolsep}{2pt}
\renewcommand{\arraystretch}{1.08}
\begin{tabular}{l|cc|cc}
\hline
\multirow{2}{*}{\textbf{Dataset}} 
& \multicolumn{2}{c|}{\textbf{Back-facing}} 
& \multicolumn{2}{c}{\textbf{Front-facing}} \\
\cline{2-5}
& Ref Acc & Cam Agree & Ref Acc & Cam Agree \\
\hline
Spatial-MM   & 0.625 & 0.250 & 0.242 & 0.758 \\
ViewSpatial  & 0.585 & 0.415 & 0.380 & 0.620 \\
3DSRBench    & 0.482 & 0.518 & 0.457 & 0.543 \\
\hline
\end{tabular}
\caption{Orientation-conditioned analysis on the conflict subset.}
\label{tab:angle_cross_dataset}
\end{table}

We further analyze how the reference object's orientation affects shortcut behavior. Table~\ref{tab:angle_cross_dataset} reports the results on the conflict subset under two coarse orientation regimes: back-facing and front-facing. Overall, front-facing conflict cases generally show stronger camera-centric behavior than back-facing ones. This trend is most evident on Spatial-MM and ViewSpatial, where camera agreement increases while reference accuracy decreases in the front-facing regime. A similar but weaker pattern is also observed on 3DSRBench, suggesting that the same directional effect exists, but with a smaller magnitude.

These results support the conclusion that object orientation is an important factor underlying camera-centric shortcut behavior. More specifically, the shortcut becomes more pronounced when the reference object's facing direction is more aligned with the image plane, making it easier for the model to fall back on camera-view cues instead of performing the required reference-frame transformation. Combined with the analysis in Section~3, this finding further suggests that the key missing component in current MLLMs is not generic spatial perception, but explicit orientation-aware modeling for reference-centric reasoning.

\section{Conclusion}

In this paper, we study perspective-aware spatial reasoning in multimodal large language models and show that current models still suffer from a strong camera-centric shortcut, especially when camera-centric and reference-centric answers conflict. Our analysis further reveals that this failure is closely related to the orientation of the reference object, suggesting that the lack of explicit orientation modeling is a key bottleneck for reference-centric reasoning.
To address this issue, we propose OrientSAM, an orientation-aware spatial alignment framework that explicitly injects object orientation into multimodal representations and aligns it with downstream spatial reasoning. By combining multi-expert spatial data construction, Fourier-based angle encoding, and curriculum learning, OrientSAM enables the model to better establish local reference frames and perform perspective-aware reasoning.
Experiments on Spatial-MM, ViewSpatial, and 3DSRBench demonstrate that OrientSAM consistently improves over strong baselines, with particularly clear gains on non-camera-view, person-centric, and orientation-sensitive tasks. These results highlight the importance of explicit orientation modeling for mitigating camera-centric bias and enabling more robust allocentric spatial reasoning in multimodal models.

\bibliography{orientsam}

\appendix

\section{Camera-Centric Shortcut Analysis Details}

To further clarify the diagnostic framework used in Section~3, we provide additional details on how camera-centric shortcut behavior is defined and measured. Our analysis focuses on perspective-taking questions that ask for the spatial relation between two entities from the viewpoint of a reference object. The key question is whether a model truly performs reference-frame transformation, or instead follows a simpler camera-centric heuristic.

For each sample, we define two candidate answers. The first is the \emph{reference-centric answer}, denoted by $\mathrm{Ans}_{\mathrm{ref}}$, which is the ground-truth spatial relation under the local viewpoint of the reference object. The second is the \emph{camera-centric answer}, denoted by $\mathrm{Ans}_{\mathrm{cam}}$, which is the answer implied by the horizontal image-plane arrangement. In our setting, the answer space is restricted to binary left/right relations. Let $x_q$ and $x_r$ denote the horizontal center coordinates of the query object and the reference object, respectively. We compute $\mathrm{Ans}_{\mathrm{cam}}$ as
\begin{equation}
\mathrm{Ans}_{\mathrm{cam}} =
\begin{cases}
\texttt{left}, & x_q < x_r,\\
\texttt{right}, & x_q > x_r.
\end{cases}
\end{equation}
This definition isolates the image-plane bias and allows us to compare it against the true reference-dependent answer.

Based on whether these two answers agree, we divide the samples into two subsets. A sample is called \emph{consistent} if
\begin{equation}
\mathrm{Ans}_{\mathrm{cam}} = \mathrm{Ans}_{\mathrm{ref}},
\end{equation}
and \emph{conflict} otherwise:
\begin{equation}
\mathrm{Ans}_{\mathrm{cam}} \neq \mathrm{Ans}_{\mathrm{ref}}.
\end{equation}
This split is central to the analysis. On consistent samples, the camera-centric answer and the reference-centric answer coincide, so a correct prediction does not necessarily indicate genuine perspective-taking. In contrast, conflict samples explicitly require the model to depart from the image-plane cue and reason under the reference object's local frame. Therefore, shortcut behavior becomes most visible on the conflict subset.

We evaluate model behavior from two complementary perspectives. The first is \emph{reference-centric accuracy}, which measures whether the model prediction $\hat{y}$ matches the ground-truth answer:
\begin{equation}
\mathrm{RefAcc} = \mathbb{I}[\hat{y} = \mathrm{Ans}_{\mathrm{ref}}].
\end{equation}
The second is \emph{camera-centric agreement}, which measures whether the prediction follows the image-plane relation:
\begin{equation}
\mathrm{CamAgree} = \mathbb{I}[\hat{y} = \mathrm{Ans}_{\mathrm{cam}}].
\end{equation}
Importantly, accuracy is always defined with respect to $\mathrm{Ans}_{\mathrm{ref}}$, while $\mathrm{Ans}_{\mathrm{cam}}$ is introduced only as a diagnostic variable. A model that truly performs reference-centric reasoning should achieve high $\mathrm{RefAcc}$ even when $\mathrm{Ans}_{\mathrm{cam}}$ disagrees with the ground truth. By contrast, if shortcut behavior exists, $\mathrm{CamAgree}$ will exceed $\mathrm{RefAcc}$ on the conflict subset.

For the model-wise analysis on ViewSpatial, we further report a normalized shortcut indicator:
\begin{equation}
\Delta_{\mathrm{norm}} = 
\frac{\mathrm{CamAgree} - \mathrm{RefAcc}}{\mathrm{RefAcc}},
\end{equation}
computed on the conflict subset. A positive $\Delta_{\mathrm{norm}}$ indicates that model predictions are more aligned with the camera-centric answer than with the reference-centric ground truth, and thus reflect a stronger shortcut tendency.

Following this framework, our main analysis is conducted on three left/right person-perspective benchmarks: Spatial-MM, ViewSpatial, and 3DSRBench. For all three datasets, each sample is converted into a unified format containing the image, the question, the reference-centric answer, the camera-centric answer derived from the image plane, the reference and query regions, and the estimated orientation of the reference object. This unified formulation allows the same diagnostic metrics to be applied across datasets.

Beyond the consistent/conflict split, we also analyze how shortcut strength varies with the orientation of the reference object. Under the angle semantics adopted in our analysis, $0^\circ$ corresponds to front-facing and $180^\circ$ corresponds to back-facing relative to the camera. To measure the distance from the back-facing condition, we define
\begin{equation}
\theta_{\mathrm{abs}} = \mathrm{circ\_dist}(\theta_{\mathrm{raw}}, 180^\circ),
\end{equation}
where $\theta_{\mathrm{raw}}$ denotes the original orientation angle and $\mathrm{circ\_dist}(\cdot,\cdot)$ is the circular distance. We then divide the samples into six intervals:
\begin{equation}
[0,30),\ [30,60),\ [60,90),\ [90,120),\ [120,150),\ [150,180].
\end{equation}
In the paper, we additionally summarize the angle effect with two coarse regimes, namely back-facing $[0,60)$ and front-facing $[120,180]$, in order to study whether camera-centric shortcut behavior changes under different reference orientations.

This diagnostic framework supports three observations in the main text. First, conflict samples are generally harder than consistent ones, indicating that many models struggle when reference-centric reasoning diverges from image-plane cues. Second, on the conflict subset, camera-centric agreement is consistently higher than reference-centric accuracy, which reveals a clear shortcut tendency. Third, the gap between these two quantities varies with the orientation of the reference object, suggesting that object orientation is an important factor underlying camera-centric shortcut behavior. In particular, the angle analysis is used as a diagnostic tool to study how shortcut strength changes across orientation conditions, rather than as an alternative definition of correctness.

\section{Multi-expert Spatial Data Construction}

To alleviate the lack of supervision for perspective-aware spatial reasoning under non-camera reference frames, we build a multi-expert spatial data construction pipeline that automatically derives spatial attributes and orientation-aware relations from large-scale images. The goal of this pipeline is to transform unstructured 2D image data into spatial question-answer pairs with explicit geometric grounding, so that the model can learn both object orientation perception and reference-frame transformation from scalable supervision.

The overall pipeline consists of five stages: image filtering, semantic tag extraction, 2D instance reconstruction, 3D geometric reconstruction, and automatic spatial annotation. Starting from large-scale image collections, we first retain only images that are suitable for constructing spatial reasoning samples. We then identify semantically meaningful entities in the image, localize them with region-level grounding, estimate their instance-level depth and orientation, and finally derive spatial relations under different reference frames. This design allows us to construct orientation-aware supervision without requiring manual 3D annotation.

In the first stage, we perform image filtering to select images with appropriate spatial structure. Since not all web images are suitable for perspective-aware reasoning, we retain only samples that satisfy several constraints, including a moderate number of salient entities, clear depth layering, recognizable semantic orientation, and relatively stable composition. In practice, we use an instruction-driven vision-language model to assess whether an image is appropriate for subsequent spatial annotation. This stage improves the overall quality of the generated data by removing images with severe ambiguity, excessive clutter, extreme camera tilt, or missing orientation cues.

In the second stage, for each retained image $I$, we extract a set of semantic tags
\begin{equation}
T = \{t_1, t_2, \dots, t_M\},
\end{equation}
where each $t_i$ denotes a semantic category that appears in the image. These semantic tags serve as textual prompts for the following grounding stage, and provide the semantic anchor for constructing object-level spatial relations.

In the third stage, we reconstruct 2D instance-level information by combining text-guided detection and segmentation. Specifically, we use GroundingDINO to obtain candidate bounding boxes conditioned on the extracted semantic tags, and then refine them with SAM to produce pixel-level masks. This yields an instance set
\begin{equation}
O = \{O_i\}_{i=1}^{n},
\end{equation}
where each instance is represented as
\begin{equation}
O_i = (b_i, m_i, t_i).
\end{equation}
Here, $b_i \in \mathbb{R}^{4}$ denotes the normalized 2D bounding box, $m_i$ is the corresponding segmentation mask, and $t_i \in T$ is the semantic label. Since a single semantic tag may correspond to multiple object instances, the number of instances $n$ is not necessarily equal to $M$.

In the fourth stage, we reconstruct 3D-related attributes from the monocular image. Given an input image $I$, the monocular geometry model predicts
\begin{equation}
(D_{\mathrm{img}}, K, \xi) = F_{\mathrm{geo}}(I),
\end{equation}
where $D_{\mathrm{img}} \in \mathbb{R}^{H \times W}$ is the dense depth map, $K \in \mathbb{R}^{3 \times 3}$ is the estimated camera intrinsic matrix, and $\xi \in SE(3)$ denotes the camera pose. Based on the segmentation mask of each instance, we further compute an instance-level depth representation. To reduce the influence of local noise and outliers, we use the median depth inside the mask as the depth of instance $O_i$:
\begin{equation}
z_i = \mathrm{Median}\left(\left\{D_{\mathrm{img}}(u,v)\mid (u,v)\in m_i\right\}\right).
\end{equation}

Besides depth, object orientation is another key attribute in our pipeline. For each instance region cropped from the image, we apply an orientation estimation model to obtain
\begin{equation}
(\hat{\theta}_i, \hat{\phi}_i, \hat{\delta}_i, c_i) = F_{\mathrm{ori}}(\mathrm{Crop}(I, b_i)),
\end{equation}
where $\hat{\theta}_i$, $\hat{\phi}_i$, and $\hat{\delta}_i$ denote the predicted orientation parameters, and $c_i \in [0,1]$ is the confidence score. The confidence score is used as a data quality signal to filter out samples whose semantic front direction is ambiguous or unreliable. In this way, the constructed data emphasize instances with stable and interpretable orientation semantics.

After the above steps, each entity is represented by a unified spatial attribute tuple:
\begin{equation}
V_i = \langle t_i, b_i, m_i, z_i, \hat{\theta}_i, \hat{\phi}_i, \hat{\delta}_i, c_i \rangle.
\end{equation}
This representation combines semantic category, 2D localization, depth, and orientation, and serves as the basis for subsequent relation derivation.

In the final stage, we automatically generate orientation-aware spatial annotations. For a pair of entities, one is treated as the reference object and the other as the query object. Using their image-plane positions, instance-level depth, and the orientation of the reference object, we derive the relative relation under the reference-centered local frame. Compared with standard camera-view labeling, this step explicitly introduces reference-frame transformation into the annotation process. Therefore, the generated supervision is not limited to image-plane left/right relations, but instead reflects the spatial relation from the viewpoint of the reference object.

Based on the derived spatial attributes and relations, we construct spatial question-answer pairs for instruction tuning. These samples cover both basic orientation-related perception and more challenging perspective-aware reasoning under non-camera reference frames. In particular, the generated data are used together with the curriculum learning strategy described in the main text, so that the model can gradually learn from simpler orientation perception tasks to harder reference-frame transformation tasks.

The proposed multi-expert pipeline offers two main advantages. First, it avoids the need for expensive manual 3D annotation while still providing explicit supervision for orientation-sensitive spatial reasoning. Second, by integrating depth estimation, region grounding, and semantic orientation prediction into a unified process, it produces training data that are more directly aligned with the target reasoning behavior of our model. Although the quality of the generated annotations is still influenced by the accuracy of the expert models, this pipeline provides an effective and scalable way to construct orientation-aware spatial supervision from large-scale 2D images.

\section{Orientation-aware Representation and Training Details}

A central goal of OrientSAM is to explicitly inject object orientation into the multimodal representation space, so that spatial reasoning can be performed under the local viewpoint of a reference object rather than the default camera plane. To this end, we augment the input representation with orientation-aware tokens and angle embeddings, and train the model with a progressive curriculum from basic orientation perception to perspective-aware spatial reasoning.

Given an image and a set of grounded regions, each object instance is associated not only with its semantic category and region information, but also with an estimated orientation. In our formulation, orientation is treated as a structured spatial attribute rather than a purely textual cue. Therefore, instead of only describing orientation in natural language, we explicitly encode it into the multimodal input sequence, allowing the model to access orientation priors at the representation level.

Concretely, for each target region, we introduce a dedicated orientation-aware placeholder token to indicate that the corresponding object is associated with orientation information. This token is used together with other region-related spatial markers in the input sequence, so that the model can distinguish ordinary semantic entities from entities carrying explicit geometric cues. In this way, orientation becomes part of the model's internal multimodal representation, rather than remaining an external textual hint.

To represent the orientation angle, we adopt Fourier-based angle encoding. A direct scalar encoding of angles suffers from discontinuity at the periodic boundary, where numerically distant values may correspond to geometrically similar directions. This issue is particularly undesirable for spatial reasoning, since nearby orientations should induce smooth changes in the representation space. To address this, we map the angle variable into a continuous high-dimensional embedding using multi-frequency Fourier features. For an angle $\alpha$, the encoding is written as
\begin{equation}
\gamma(\alpha)=
\left[
\begin{aligned}
&\sin(2^{0}\pi\alpha),\ \cos(2^{0}\pi\alpha),\\
&\sin(2^{1}\pi\alpha),\ \cos(2^{1}\pi\alpha),\\
&\ldots,\\
&\sin(2^{L-1}\pi\alpha),\ \cos(2^{L-1}\pi\alpha)
\end{aligned}
\right].
\end{equation}
where $L$ denotes the number of frequency bands. This mapping preserves the periodic structure of the angle variable and improves the smoothness and continuity of orientation-aware representations.

The resulting Fourier embedding is fused with the visual region representation, so that orientation information can participate in cross-modal alignment and downstream reasoning. Instead of treating angle as a separate symbolic label, this design enables the model to associate object appearance, spatial region, and orientation within a unified feature space. As a result, the model is encouraged to learn orientation-sensitive spatial patterns directly from representation interaction.

Based on this orientation-aware representation, the training process is further organized with a curriculum learning strategy. The motivation is that perspective-aware reasoning is a compositional ability: the model must first perceive the orientation of the reference object, and then use this information to transform the spatial relation into the corresponding local frame. Directly training on the final reasoning task alone makes this learning process unstable. Therefore, we adopt a progressive training schedule that gradually increases task difficulty.

At the early stage, the model is trained on simpler orientation-related perception tasks, which mainly strengthen its sensitivity to semantic front direction and orientation-dependent object understanding. This stage helps the model establish an initial correspondence between visual appearance and orientation embedding. After this basic capability is formed, the training data are gradually shifted toward more complex spatial reasoning samples that require reference-frame transformation. In these tasks, the model must combine region grounding, depth-related geometry, and the orientation of the reference object to infer the correct relation under a non-camera viewpoint.

This progressive strategy reduces the optimization difficulty of learning perspective-aware reasoning from scratch. In particular, it encourages the model to acquire orientation perception first, and then reuse this ability as a geometric prior for higher-level spatial reasoning. Such a staged training process is especially important in our setting, because the target reasoning behavior depends on both low-level direction understanding and high-level reference-frame transformation.

Overall, the orientation-aware representation and training design in OrientSAM consists of two tightly coupled components. The first is an explicit orientation-aware representation that encodes angle information through dedicated tokens and Fourier-based embeddings. The second is a curriculum learning schedule that organizes training from basic orientation perception to complex perspective-aware reasoning. Together, these two components allow orientation priors to be integrated into the multimodal hidden space in a smooth and learnable manner, thereby improving the model's robustness on orientation-sensitive spatial reasoning tasks.

\section{Additional Experimental Settings}

We provide additional experimental details to improve clarity and reproducibility. Our experiments focus on perspective-aware spatial reasoning under non-camera reference frames, where the model is required to infer spatial relations from the local viewpoint of a reference object rather than from the image plane.

We evaluate OrientSAM on three representative spatial reasoning benchmarks, namely Spatial-MM, ViewSpatial, and 3DSRBench. These benchmarks are all used to assess whether multimodal models can correctly perform reference-frame transformation in left/right spatial reasoning tasks. For analysis purposes, all samples are converted into a unified format containing the image, the question, the ground-truth reference-centric answer, the corresponding camera-centric answer derived from image-plane geometry, the reference and query regions, and the estimated orientation of the reference object. This unified formulation allows us to conduct consistent shortcut analysis across different datasets.

Following the setting in the main paper, the evaluation mainly focuses on binary left/right reasoning under a reference object's viewpoint. In addition to standard answer accuracy, we further analyze model behavior using the diagnostic framework introduced in Section~3. Specifically, for each sample, we compare the model prediction with both the reference-centric answer and the camera-centric answer, and divide the benchmark into consistent and conflict subsets according to whether these two answers agree. This protocol is used throughout our shortcut analysis and orientation-based analysis.

For model comparison, we consider both strong multimodal baselines and our orientation-aware model under the same evaluation setting. All compared models are prompted to answer the same spatial question without access to ground-truth orientation labels at test time. The goal is to ensure that performance differences mainly reflect differences in spatial reasoning ability rather than differences in task formulation. During evaluation, the model output is mapped to the predefined answer space, and only the final left/right decision is used for scoring.

For training data construction, we use the multi-expert spatial pipeline described in the previous section to automatically generate orientation-aware supervision from large-scale images. The resulting data contain both basic orientation-related perception samples and more complex perspective-aware reasoning samples. These samples are then organized by the curriculum learning strategy in the main paper, so that the training process gradually shifts from easier orientation perception tasks to harder reference-frame transformation tasks.

In terms of implementation, the orientation signal is incorporated into the model through explicit orientation-aware tokens and Fourier-based angle encoding. The model is trained under the same general multimodal instruction-tuning framework as the baseline architecture, while the main architectural change lies in the injection and alignment of orientation-aware spatial representations. During training, the curriculum is applied progressively, so that the model first learns stable orientation perception and then adapts this capability to downstream spatial reasoning tasks involving non-camera viewpoints.

For the orientation-based analysis, we follow the angle semantics adopted in our shortcut study. In particular, the reference object's orientation is represented relative to the camera, and the samples are grouped by angle intervals to examine how model behavior changes with orientation. This analysis is used only as a diagnostic tool for understanding shortcut strength and orientation sensitivity; it does not change the definition of the ground-truth answer.

Overall, our experimental setting is designed to evaluate not only whether a model can produce the correct answer, but also whether the answer is obtained through genuine reference-centric reasoning rather than by relying on image-plane shortcuts. Therefore, besides benchmark accuracy, we emphasize consistent/conflict analysis, camera-agreement analysis, and orientation-based behavior analysis as complementary evaluation dimensions.


\section{Robustness to Upstream Expert Errors}

OrientSAM relies on grounding, depth, and orientation experts when constructing its offline training supervision. To measure how errors from these stages affect downstream reasoning, we independently perturb one expert output at a time while keeping the trained model and evaluation benchmark fixed. Grounding boxes are shifted in center and size, depth estimates are shifted across ordered bins or inverted, and orientation estimates are shifted across 5-degree yaw bins. Table~\ref{tab:propagation} reports the resulting accuracy on 3DSRBench.

\begin{table*}[t]
\centering
\setlength{\tabcolsep}{8pt}
\begin{tabular}{lll}
\toprule
\textbf{Protocol} & \textbf{Perturbation Strength} & \textbf{Accuracy} \\
\midrule
Clean pipeline & None & 76.0 \\
Grounding: box center/size shift & $\pm5\% / \pm10\% / \pm20\%$ of box scale & 73.8 / 72.2 / 67.4 \\
Depth: ordered-bin shift & $\pm1 / \pm3$ bins; inversion & 72.1 / 67.3 / 62.6 \\
Orientation: yaw-bin shift & $\pm1 / \pm2 / \pm10$ in 5-degree bins & 74.7 / 72.4 / 62.7 \\
\bottomrule
\end{tabular}
\caption{Cascaded-error stress test on 3DSRBench. Each perturbation is applied to one upstream expert stage at a time.}
\label{tab:propagation}
\end{table*}

Small perturbations cause moderate degradation, whereas severe corruptions lead to larger drops. This result indicates that OrientSAM tolerates moderate upstream noise but still inherits an expected failure boundary from its expert-generated supervision. During data construction, low-confidence or orientation-ambiguous outputs are therefore filtered rather than forced into the training set.

\section{Data Independence and Gain Attribution}

The training images are drawn from general large-scale image collections, including BLIP3-Kale and LAION, rather than from Spatial-MM, ViewSpatial, or 3DSRBench images and test question-answer pairs. The 550K-example training set contains 150K constructed samples---50K orientation-perception samples and 100K non-camera-view reasoning samples---together with 400K auxiliary spatial/VQA examples. We retain orientation annotations with confidence $c_i>0.98$. A manual audit of 200 non-camera-view examples found 93\% to have correct reference/target grounding and consistent reference-frame question-answer labels.

We additionally apply source and image-ID filtering followed by CLIP nearest-neighbor de-duplication. We then retrain the curriculum stage using the filtered data. As shown in Table~\ref{tab:dedup}, performance remains close to the original result across all 3DSRBench subsets, suggesting that the observed gains are unlikely to be explained by near-duplicate overlap.

\begin{table*}[t]
\centering
\setlength{\tabcolsep}{8pt}
\begin{tabular}{lcccc}
\toprule
\textbf{Setting} & \textbf{Height} & \textbf{Location/Distance} & \textbf{Orientation/Direction} & \textbf{Multi-object} \\
\midrule
OrientSAM & 56.67 & 78.23 & 81.83 & 77.80 \\
De-duplicated data + retrained curriculum & 56.32 & 78.56 & 80.13 & 76.47 \\
\bottomrule
\end{tabular}
\caption{Results after CLIP nearest-neighbor de-duplication and curriculum-stage retraining on 3DSRBench.}
\label{tab:dedup}
\end{table*}

To separate orientation-aware alignment from a data-only explanation, we further conduct two counterfactual controls under fixed data and compute. The first verbalizes either the orientation angle or the induced front/back/left/right frame in an SRGPT prompt. The second keeps the image, question-answer pair, grounding, depth, training schedule, and token structure unchanged, but replaces each orientation token with one sampled from another example in the same object-category group.

\begin{table}[t]
\centering
\setlength{\tabcolsep}{3pt}
\begin{tabular}{lc}
\toprule
\textbf{Setting} & \textbf{Non-Cam Acc.} \\
\midrule
Same-data SFT & 43.15 \\
SRGPT + text-angle prompt & 41.23 \\
SRGPT + text-frame prompt & 46.38 \\
OrientSAM with shuffled orient. & 66.37 \\
Full OrientSAM & \textbf{87.31} \\
\bottomrule
\end{tabular}
\caption{Counterfactual controls on Spatial-MM Non-Cam.}
\label{tab:counterfactual}
\end{table}

Table~\ref{tab:counterfactual} shows that textual orientation prompts alone do not reproduce the full improvement. Shuffling the orientation signal also reduces accuracy by 20.94 points relative to full OrientSAM. This gap supports the contribution of correct, instance-aligned orientation information, while the remaining improvement over same-data SFT suggests that other structured spatial cues also contribute.

\section{Efficiency and External Benchmark Evaluation}

Table~\ref{tab:inference} compares inference time under the same single-A100 setting on Spatial-MM non-camera-view examples. OrientSAM introduces additional computation relative to the vanilla Qwen2.5-VL-7B backbone and SRGPT, but remains faster than APC-Vis while achieving slightly higher accuracy on this orientation-sensitive subset.

\begin{table}[t]
\centering
\setlength{\tabcolsep}{5pt}
\begin{tabular}{lcc}
\toprule
\textbf{Model} & \textbf{Seconds/sample} & \textbf{Accuracy} \\
\midrule
Qwen2.5-VL-7B & \textbf{6.13} & 37.31 \\
SRGPT & \underline{10.21} & 39.55 \\
APC-Vis & 18.13 & \underline{86.57} \\
OrientSAM & 12.32 & \textbf{87.31} \\
\bottomrule
\end{tabular}
\caption{Single-A100 inference on Spatial-MM Non-Cam.}
\label{tab:inference}
\end{table}

We also test transfer on OmniSpatial's three perspective-taking splits and MMSI-Bench. Table~\ref{tab:recent_benchmarks} shows the clearest gain on OmniSpatial allocentric reasoning, from 33.2 to 44.6, with smaller gains elsewhere.

\begin{table}[t]
\centering
\setlength{\tabcolsep}{3pt}
\begin{tabular}{lcccc}
\toprule
\multirow{2}{*}{\textbf{Model}} & \multicolumn{3}{c}{\textbf{OmniSpatial}} & \multirow{2}{*}{\textbf{MMSI}} \\
\cmidrule(lr){2-4}
& \textbf{Ego} & \textbf{Allo} & \textbf{Hypo} & \\
\midrule
Qwen2.5-VL-7B & 64.5 & 33.2 & 37.4 & 25.9 \\
OrientSAM & 67.1 & 44.6 & 43.9 & 27.0 \\
\bottomrule
\end{tabular}
\caption{Additional spatial benchmark results.}
\label{tab:recent_benchmarks}
\end{table}

Together, these results calibrate the method's scope: OrientSAM targets reference-frame transformation, with its strongest gains on direction-sensitive, non-camera-view, and allocentric reasoning rather than generic visual perception.

\section{Limitations and Discussion}

Although OrientSAM improves perspective-aware spatial reasoning by explicitly incorporating orientation information, several limitations remain. First, the overall framework still relies on the quality of external expert models in the data construction stage, including region grounding, monocular geometry estimation, and object orientation prediction. Errors from these components may propagate to the automatically generated supervision, which in turn affects the quality of the learned spatial reasoning behavior.

Second, the effectiveness of the proposed approach depends on whether the reference object has a stable and semantically meaningful front direction. In real images, some objects have ambiguous orientation, weak visual cues, or heavy occlusion, making their semantic facing direction difficult to estimate reliably. In such cases, even explicit orientation modeling may provide limited benefit, since the reference frame itself is not well defined.

Third, our current study mainly focuses on left/right spatial reasoning under a reference object's viewpoint. While this setting is sufficient for diagnosing camera-centric shortcut behavior and validating the importance of orientation-aware modeling, it only covers a subset of perspective-aware spatial reasoning. More complex relations, such as front/behind, relative distance, and multi-object reasoning under dynamic reference frames, are not fully explored in the current work.

Despite these limitations, our results suggest that object orientation is an important missing factor in current multimodal spatial reasoning systems. The analysis in the main paper shows that existing models often default to camera-plane heuristics when reference-centric reasoning conflicts with image-plane cues, and that this shortcut behavior is closely related to the orientation of the reference object. From this perspective, our method should be viewed not only as a performance-oriented improvement, but also as an attempt to introduce a more appropriate spatial inductive bias into multimodal representations.

More broadly, the proposed framework highlights the importance of explicitly modeling local reference frames in multimodal reasoning. Many practical scenarios, such as human-object interaction and scene understanding, require the model to reason from the viewpoint of an entity in the scene rather than from the camera alone. Our findings indicate that enhancing multimodal models with orientation-aware representations and progressively trained perspective transformation ability is a promising step toward more robust allocentric spatial reasoning.

In future work, it would be valuable to reduce the dependence on external expert annotations, extend the formulation to richer spatial relation types, and study how orientation-aware reasoning can be integrated into broader embodied or interactive multimodal settings.

\end{document}